\documentclass{article}

\PassOptionsToPackage{square,numbers,sort&compress}{natbib}
\usepackage[accepted]{icml2021}

\usepackage{amsmath,amsfonts,bm}

\def\eqref#1{equation~\ref{#1}}

\def\1{\bm{1}}

\DeclareMathAlphabet{\mathsfit}{\encodingdefault}{\sfdefault}{m}{sl}
\SetMathAlphabet{\mathsfit}{bold}{\encodingdefault}{\sfdefault}{bx}{n}

\usepackage[utf8]{inputenc} %
\usepackage[T1]{fontenc}    %
\usepackage[pagebackref=true,breaklinks=true,colorlinks,bookmarks=false]{hyperref}
\usepackage{url}            %
\usepackage{booktabs}       %
\usepackage{amsfonts}       %
\usepackage{nicefrac}       %
\usepackage{microtype}      %

\usepackage{dblfloatfix}
\usepackage{graphicx}
\usepackage{color}
\usepackage{caption}
\usepackage{enumitem}
\usepackage{tabularx}
\usepackage{colortbl}  %
\usepackage{bm}  %
\usepackage{amssymb}
\usepackage{amsmath}
\usepackage{wasysym}
\usepackage{wrapfig}
\usepackage{placeins}
\usepackage{subfig}

\definecolor{darkergreen}{rgb}{0.0, 0.5, 0.0}

\definecolor{Gray}{gray}{0.9}
\newcolumntype{g}{>{\columncolor{Gray}}c}  %
\newcolumntype{H}{>{\setbox0=\hbox\bgroup}c<{\egroup}@{}}  %
\definecolor{GrayLine}{gray}{0.7}
\newcolumntype{Y}{>{\centering\arraybackslash}X}

\definecolor{demphcolor}{RGB}{100,100,100}
\newcommand{\demph}[1]{\textcolor{demphcolor}{#1}}
\newcommand{\pms}[2]{#1{\tiny{{\demph{{$\pm$#2}}}}}}
\newcolumntype{x}[1]{>{\centering\arraybackslash}p{#1pt}}
\newlength\savewidth
\newcommand{\tablestyle}[2]{\setlength{\tabcolsep}{#1}\renewcommand{\arraystretch}{#2}\centering\footnotesize}
\newcommand{\colw}{30}

\newcommand{\bb}{\mathbf{b}}
\newcommand{\be}{\mathbf{e}}
\newcommand{\bj}{\mathbf{j}}
\newcommand{\bp}{\mathbf{p}}
\newcommand{\bu}{\mathbf{u}}
\newcommand{\by}{\mathbf{y}}
\newcommand{\bw}{\mathbf{w}}
\newcommand{\bW}{\mathbf{W}}
\newcommand{\bz}{\mathbf{z}}

\icmltitlerunning{Physical Reasoning Using Dynamics-Aware Models}

\begin{document}

\twocolumn[
\icmltitle{Physical Reasoning Using Dynamics-Aware Models}

\icmlsetsymbol{equal}{*}
\begin{icmlauthorlist}
    \icmlauthor{Eltayeb Ahmed}{ed}
    \icmlauthor{Anton Bakhtin}{ed}
    \icmlauthor{Laurens van der Maaten}{ed}
    \icmlauthor{Rohit Girdhar}{ed}
\end{icmlauthorlist}
\center{\small \url{https://facebookresearch.github.io/DynamicsAware}}

\icmlaffiliation{ed}{Facebook AI Research, New York}

\icmlcorrespondingauthor{Eltayeb Ahmed}{eltayeb.k.ahmed@gmail.com}
\icmlcorrespondingauthor{Rohit Girdhar}{rgirdhar@fb.com}

\icmlkeywords{Physical Reasoning, Value Learner, Deep Learning}
\vskip 0.3in
]

\printAffiliationsAndNotice{}

\begin{abstract}
A common approach to solving physical-reasoning tasks is to train a value learner on example tasks.
A limitation of such an approach is that it requires learning about object dynamics solely from reward values assigned to the final state of a  rollout of the environment.
This study aims to address this limitation by augmenting the reward value with self-supervised signals about object dynamics.
Specifically, we train the model to characterize the similarity of two environment rollouts, jointly with predicting the outcome of the reasoning task.
This similarity can be defined as 
a distance measure between the trajectory of objects in the two rollouts, 
or learned directly from pixels using a contrastive formulation.
Empirically, we find that this approach leads to substantial performance improvements on the PHYRE benchmark for physical reasoning~\cite{bakhtin2019phyre}, establishing a new state-of-the-art.
\end{abstract}

\section{Introduction}
\label{sec:introduction}

Many open problems in artificial intelligence require agents to reason about physical interactions between objects.
Spurred by the release of benchmarks such as Tools~\cite{allen2019tools} and PHYRE~\cite{bakhtin2019phyre}, such \emph{physical reasoning} tasks have become a popular subject of study~\cite{girdhar2020forward,qi2021longterm,Whitney2020DynamicsawareE}.
Specifically, physical-reasoning tasks define an \emph{initial state} and a \emph{goal state} of the world, and require selecting an \emph{action} that comprises placing one or more additional objects in the world.
After the action is performed, the world simulator is unrolled to determine whether or not the goal state is attained.
Despite their simplicity, benchmarks like PHYRE are surprisingly difficult to solve due to the chaotic nature of the dynamics of physical objects. 
Current approaches for physical reasoning problems can be subdivided into two main types:
\begin{enumerate}[leftmargin=*]
\item \textbf{Dynamics-agnostic approaches} treat the problem as a ``standard'' contextual bandit that tries to learn the value of taking a particular action given an initial state, without using the simulator rollout in any way~\cite{bakhtin2019phyre}. 
An advantage of such approaches is that they facilitate the use of popular learning algorithms for this setting, such as deep Q-networks (DQNs; \citet{mnih-atari-2013}) 
However, the approaches do not use information from the simulator rollout as learning signal, which limits their efficacy.
\item \textbf{Dynamics-modeling approaches} learn models that explicitly aim to capture the dynamics of objects in the world, and use those models to perform forward prediction~\cite{girdhar2020forward,qi2021longterm,Whitney2020DynamicsawareE}.
Such forward predictions can then be used, for example, in a search algorithm to find an action that is likely to be successful.
An advantage of such approaches is that they use learning signal obtained from the simulator rollout.
However, despite recent progress~\cite{gonzalez2020learning}, high-fidelity dynamics prediction in environments like PHYRE remains an unsolved problem~\cite{girdhar2020forward}.
Moreover, current approaches do not use the uncertainty in the dynamics model to select actions that are most likely to solve the task.
\end{enumerate}
In this paper, we develop a \textbf{dynamics-aware} approach for physical reasoning that is designed to combine the strengths of the current two approaches.
Our approach incorporates information on simulator rollout into the learning signal used to train DQNs.
The resulting models outperforms all prior work on the PHYRE benchmark, including full dynamics-modeling approaches, achieving a new state-of-the-art of $86.2$ in the one-ball (1B), within-template setting. %

\section{Dynamics-Aware Deep Q-Networks}\label{sect:appr}

The basis of the model we develop for physical reasoning is a standard deep Q-network (DQN; \citet{mnih-atari-2013,bakhtin2019phyre}).
We augment the loss function used to train this model with a dynamics-aware loss function.
This allows the model-free DQN learner to explicitly incorporate dynamics of the environment at training time, without having to do accurate dynamics prediction at inference time.

Our backbone model is a ResNet~\cite{He_16} that takes an image depicting the initial scene, $s$, of the task as input. 
The action, $a$, parameterized as a $(x,y,r)$-vector\footnote{For the  PHYRE-2B tier, the action is parametrized via a $(x_1, y_1, r_1, x_2, y_2, r_2)$-vector.}, is processed by a multilayer perceptron with one hidden layer to construct an action embedding.
The action embedding is fused with the output of the third ResNet block using FiLM modulation~\cite{perez2017film}. This fused representation is input into the fourth block of the ResNet to obtain a scene-action embedding, $\be_{s, a}$. 
We score action $a$ by applying a linear layer with weights $\bw_1$ and bias $b_1$ on $\be_{s, a}$.
At training time, we evaluate this score using a logistic loss that compares it against a label, $y_{s,a}$, that indicates whether or not action $a$ solves task $s$:
\begin{gather}
    \hat{y}_{s,a} = \bw_1^\top \be_{s,a} + b_1 \label{eq:linear_cls} \\
    \mathcal{L} =  - \left(y_{s,a}  \log \left( \hat{y}_{s,a} \right)  +  (1 - y_{s,a})  \log \left( 1 - \hat{y}_{s,a} \right) \right).
\end{gather}
We study two approaches to incorporate a dynamics-aware loss that encourages scene-action embeddings $\be_{s, a}$ that lead to similar rollouts to be similar: 1) A \textbf{hand-crafted loss} that leverages the object-state information from the simulator to determine which rollouts are similar and tries to match them; and 2) A \textbf{self-supervised loss} that does not rely on object-state information but, instead, operates on a pixel-level embedding that it matches to an embedding computed from a random portion of its own rollout, contrasting against other rollouts in the batch. We now describe both in detail. 

\noindent\textbf{Hand-crafted dynamics-aware loss.} 
Given a pair of actions $(a, a')$ for task $s$,
we compute a joint embedding of the two actions $\bj_{s,a,a'}$ for that task as follows:
\begin{gather}
	\label{eqn:approach:joint_embedding}
    \bp_{s, a} = \text{MLP}(\be_{s,a}; \theta); \quad\quad\quad
    \bp_{s, a'} = \text{MLP}(\be_{s,a'}; \theta)  \\
	\bj_{s,a,a'} = \bp_{s,a} \odot \bp_{s,a'}.
\end{gather}

Herein, $\odot$ refers to a combination function: we use the outer product by default but we also experiment with element-wise products and concatenation in Section~\ref{sec:ablations}.
We pass $\bj_{s,a,a'}$ through another linear layer to predict the similarity of the two actions applied on scene $s$. 
The model is trained to minimize a loss that compares the predicted similarity to a ``ground-truth'' similarity.
Specifically, we bin the ground-truth similarity into $K$ bins and minimize the cross-entropy loss of predicting the right bin:
\begin{gather}
    \bu_{s,a,a'} = \bW_{2}^\top j_{s,a,a'} + \bb_{2} \\ 
    \mathcal{L}_{aux} = - \by_s^\top \bu_{s,a,a'} + \log\left(\sum_{\by'_s} \exp\left[ \by'^{\top}_s \bu_{s,a,a'} \right]\right).
\end{gather}
Herein, $\by_s$ is a one-hot vector of length $K$ indicating the bin in which the ground-truth similarity falls.

To measure the ground-truth similarity, $v_{a,a'}$, between two actions $a$ and $a'$ on task $s$, we run the simulator on the two scenes obtained after applying the actions.
We track all objects throughout the simulator roll-outs, and measure the Euclidean distance between each object in one roll-out and its counterpart in the other roll-out. 
This results in distance functions, $d_{a,a'}(o, t)$, for all objects $o \in \mathcal{O}$ (where $t$ represents time).
We convert the distance function into a similarity functions and aggregate all similarities over time and over all objects:
\begin{gather}
q_{a,a'}(o, t) = 1 - \frac{\min(d_{a,a'}(o, t), \alpha)}{\alpha} \\
v_{a,a'} = \frac{\sum_{o \in \mathcal{O}} \sum_{t=1}^T q_{a,a'}(o, t)}{T | \mathcal{O} |},
\end{gather}
where $\alpha$ is a hyperparameter that clips the distance at a maximum value, and $T$ is the number of time steps in the roll-out. The similarity $ v_{a,a'}$ is binned to construct $\by_s$. See Appendix~\ref{apndx:metric} for details. 

\noindent\textbf{Self-supervised Dynamics-Aware loss.}
Since the previous similarity metric relies on access to the ground truth (GT) object states through the rollout---an unreasonable assumption in most real-world scenarios---we propose a modification of the proposed auxiliary loss, by learning the distance between rollouts using a contrastive formulation. To that end, we start by representing a scene-action pair by rendering the action onto the scene and passing it through a vanilla ResNet-18 to obtain the embedding $\be_{s,a}$, instead of the FiLM approach used earlier.
This allows for a common representation space for different parts of the rollout without relying on GT object positions. This embedding is then used for two objectives. First, as before, we predict whether the task will be solved or not using a linear layer as in Equation~\ref{eq:linear_cls}. Second, we embed $K$ consecutive frames from a random point in the rollout using the same encoder to obtain $\be_{s,a}^1 \cdots \be_{s,a}^K$, concatenate these embeddings and pass the result through an MLP to reduce it to the same dimensions as $\be_{s,a}$. We refer to the resulting ``rollout embedding'' as $\be_{s,a}^{roll}$. We then incur a contrastive loss~\cite{oord2018representation} to match the initial embedding action embedding to its corresponding rollout embedding after passing them through another MLP:
\begin{gather}
    \bz_{s,a} = \text{MLP}(\be_{s,a}) \\ %
    \mathcal{L}_{aux} = - \beta \bz_{s,a}^T \be_{s,a}^{roll} + \log \left(\sum \exp \left( \beta \bz_{s,a}^T \be_{s,a}^{roll'} \right) \right),\nonumber
\end{gather}
with $\be_{s,a}^{roll'}$ the rollout embeddings, and $\beta$ the temperature.

\begin{table}[t]
    \setlength{\tabcolsep}{2pt}
    \footnotesize
    \centering

    \vspace{1mm}
    \tablestyle{4pt}{1.2}
    \resizebox{\linewidth}{!}{%
    \begin{tabular}{@{}lx{\colw}x{\colw}x{\colw}x{\colw}@{}}  %
        \toprule
        &  \multicolumn{2}{c}{\bf PHYRE-1B} & \multicolumn{2}{c}{\bf PHYRE-2B} \\ %
        &  \bf Within & \bf Cross & \bf Within & \bf Cross \\ %
        \midrule
        RAND~\citep{bakhtin2019phyre} & \pms{13.7}{0.5} & \pms{13.0}{5.0} & \pms{3.6}{0.6} & \pms{2.6}{1.5} \\ %
        MEM~\citep{bakhtin2019phyre} & \pms{2.4}{0.3} & \pms{18.5}{5.1} & \pms{3.2}{0.2} & \pms{3.7}{2.3} \\ %
        DQN~\citep{bakhtin2019phyre} & \pms{77.6}{1.1} & \pms{36.8}{9.7} & \pms{67.8}{1.5} & \pms{23.2}{9.1} \\ %
        Dec [Joint]~\cite{girdhar2020forward}&  \pms{80.0}{1.2} & \pms{40.3}{8.0} &  -- &  -- \\ %
        RPIN~\cite{qi2021longterm}&  \pms{85.2}{0.7} & {\bf \pms{42.2}{7.1}} &  -- &  -- \\ %
        {\bf Ours (Hand-crafted)} & \pms{85.0}{1.1} & \pms{38.6}{8.4} &  \pms{74.0}{1.5} & \pms{23.3}{8.8} \\ %
        {\bf Ours (Self-supervised)} & {\bf \pms{86.2}{0.9}} & \pms{41.9}{8.8} & {\bf \pms{77.6}{1.4}} & {\bf \pms{24.3}{10.0}} \\
        \arrayrulecolor{black}
        \bottomrule
    \end{tabular}%
    }
        \captionof{table}{AUCCESS of dynamics-aware DQNs compared to state-of-the-art models on the PHYRE-1B and 2B tiers in the within-template and cross-template generalization settings. Results reported are averaged over 10 test folds.
    }\label{tab:expts:sota}
    \vspace{-4mm}
 \end{table}

\newcommand\pageSplit{0.50}
\newcommand\pageSplitR{0.40}

\begin{table}[t]
    \centering
    \vspace{3mm}
    \setlength{\tabcolsep}{2pt}
    \footnotesize
    \centering
    \tablestyle{4pt}{1.2}
    \resizebox{0.4\linewidth}{!}{%
    \begin{minipage}[t]{\pageSplit\linewidth}
        \centering
        \subfloat[Auxiliary loss.\label{tab:expts:ablations:backbones}]
            {\begin{tabular}{@{}lx{\colw}x{\colw}x{\colw}x{\colw}}
                \toprule
                \bf Model & \bf Within & \bf Cross \\ %
                \midrule
                DQN & \pms{81.7}{0.6} & \pms{36.4}{11.1}  \\
                \underline{Ours} &  \pms{83.1}{1.0} &  \pms{36.3}{9.3} \\
                \bottomrule
            \end{tabular}}
            \hfill
            \subfloat[Projection layer. \label{tab:expts:ablations:embeddor}]{\begin{tabular}{lc}
                \toprule
                & \bf AUCCESS \\
                \midrule
                None & \pms{82.9}{0.5} \\
                \underline{Linear}  &\pms{83.0}{1.0} \\
                2-Layer MLP &\pms{82.6}{0.9} \\
                3-Layer MLP & \pms{82.8}{1.3} \\
                \bottomrule
            \end{tabular}}
        \hfill
        \subfloat[Number of bins. \label{tab:expts:ablations:bins}]
            {\begin{tabular}{lc}
                \toprule
                $\boldsymbol{k}$  &\bf AUCCESS \\
                \midrule
                2  & \pms{81.9}{0.9}\\
                5  & \pms{82.8}{0.6} \\
                10 & \pms{82.8}{1.3} \\
                \underline{20}  & \pms{83.0}{0.6} \\
                MSE & \pms{81.4}{0.5} \\
                \bottomrule
            \end{tabular}}
    \end{minipage}}
    \hfill
    \resizebox{0.35\linewidth}{!}{%
    \begin{minipage}[t]{\pageSplitR\linewidth}
        \centering
        \subfloat[Combination function.\label{tab:expts:ablations:combination}]
        {\begin{tabular}{lc}
            \toprule
            & \bf AUCCESS \\
            \midrule
            Multiplication & \pms{82.6}{1.4}   \\
            Concatenation & \pms{81.8}{1.4}  \\
            \underline{Bilinear} & \pms{83.3}{1.1} \\
            \bottomrule
        \end{tabular}}
        \hfill
        \subfloat[Frames used in $v_{a,a'}$. \label{tab:expts:ablations:time}]{\begin{tabular}{lc}
            \toprule
            \bf Frames & \bf AUCCESS \\
            \midrule
            First 1 & \pms{81.4}{0.7} \\
            First 3 & \pms{81.9}{1.1} \\
            First 5 & \pms{82.7}{1.4} \\
            First 10 &\pms{82.9}{0.9} \\
            \arrayrulecolor{GrayLine}
            \midrule
            Last 1 & \pms{82.8}{1.0} \\
            Last 3 & \pms{82.6}{1.0} \\
            Last 5 & \pms{82.6}{1.0} \\
            Last 10 & \pms{83.0}{0.9} \\
            \arrayrulecolor{GrayLine}
            \midrule
            \underline{Entire Rollout} & \pms{83.0}{1.0} \\
            \arrayrulecolor{black}
            \bottomrule
        \end{tabular}}
    \end{minipage}}
    \hspace{0.3in}
    \caption{
        {\bf Hand-crafted loss ablations.} AUCCESS on PHYRE-1B validation folds in within-template setting.
    \label{tab:expts:ablations}}
    \vspace{-4mm}
\end{table}

\newcommand\widthFrac{0.15}
\begin{table*}[t]
    \centering
    \setlength\tabcolsep{1.5pt}
    \resizebox{\linewidth}{!}{%
    \begin{tabular}{ccc@{\hspace{1em}}ccc@{\hspace{1em}}cc}
        GT & Baseline & Ours & GT & Baseline & Ours & Baseline & Ours \\
        \frame{\includegraphics[width=\widthFrac\linewidth]{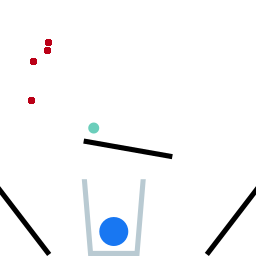}} &
        \frame{\includegraphics[width=\widthFrac\linewidth]{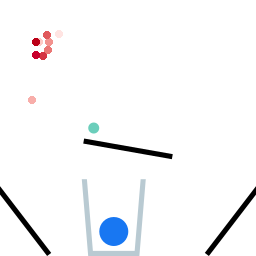}} &
        \frame{\includegraphics[width=\widthFrac\linewidth]{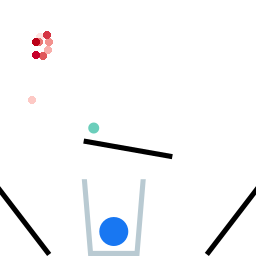}} &
        \frame{\includegraphics[width=\widthFrac\linewidth]{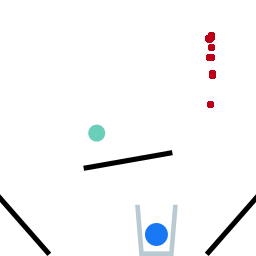}} &
        \frame{\includegraphics[width=\widthFrac\linewidth]{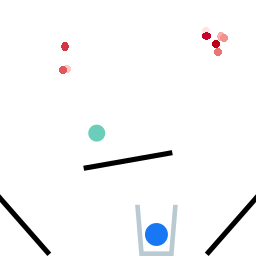}}  &\frame{\includegraphics[width=\widthFrac\linewidth]{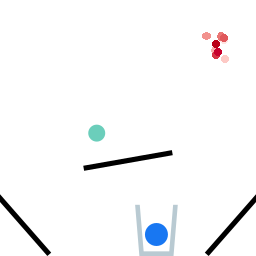}} &
        \frame{\includegraphics[width=\widthFrac\linewidth]{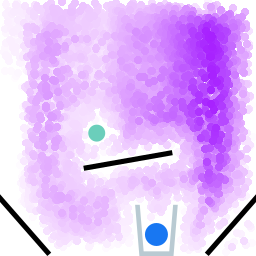}}  &
        \frame{\includegraphics[width=\widthFrac\linewidth]{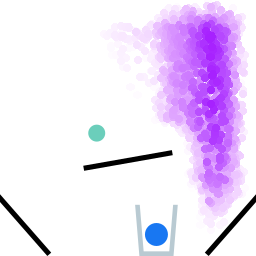}}
        \\
        \multicolumn{3}{c@{\hspace{1em}}}{ (a) Task A} & \multicolumn{3}{c@{\hspace{1em}}}{(b) Task B} & \multicolumn{2}{c}{(c) Task B top actions space} \\
    \end{tabular}}
    \captionof{figure}{In (a) and (b), we visualize all the ground truth (GT) and top 10 predicted actions' positions ($x,y$) that solves the above two tasks, with darker color representing higher confidence. On Task A, our method (hand-crafted) performs similarly to a dynamics-agnostic baseline. In Task B where the incline is slanted the other way, however, the baseline model is confused between two possible sets of positions of the action. By contrast, our dynamics-aware DQN model is able to solve this task correctly. In (c), we visualize all actions with $>0.9$ cosine similarity to any action that solves the task (darker color $\!\implies\!$ higher similarity). 
    Our dynamics-aware DQN model rules out more incorrect actions than the baseline DQN.
    }\label{fig:expts:inclination}
     \vspace{-2mm}
\end{table*}
\renewcommand\pageSplit{0.50}
\renewcommand\pageSplitR{0.40}

\begin{table}[t]
    \vspace{-1em}
    \centering
    \vspace{3mm}
    \setlength{\tabcolsep}{2pt}
    \footnotesize
    \centering
    \tablestyle{4pt}{1.2}
    \resizebox{0.45\linewidth}{!}{%
    \begin{minipage}[t]{\pageSplit\linewidth}
        \centering
        \subfloat[Action representation.\label{tab:expts:c_ablations:action_rep}]
            {\begin{tabular}{lcc}
                \toprule
                \bf Model & \bf AUCCESS \\ %
                \midrule
                DQN &  \pms{83.0}{0.6}  \\
                \underline{Ours} & \pms{84.7}{1.3}\\
                \bottomrule
            \end{tabular}}
            \hfill
            \subfloat[Contr. loss weight. \label{tab:expts:c_ablations:weight}]
            {\begin{tabular}{lc}
                \toprule
                \bf Weight  &\bf AUCCESS \\
                \midrule
                0.01  & \pms{82.8}{1.2}\\
                0.05  & \pms{82.9}{1.1} \\
                0.1 & \pms{82.9}{0.9} \\
                0.5 & \pms{83.0}{0.7} \\
                \underline{1} & \pms{83.7}{1.0} \\
                5 & \pms{82.6}{1.0} \\
                \bottomrule
            \end{tabular}}
        \hfill
        \subfloat[Projection layer. \label{tab:expts:c_ablations:proj}]{\begin{tabular}{lc}
            \toprule
            & \bf AUCCESS \\
            \midrule
            None & \pms{81.8}{1.3} \\
            \underline{Linear}  &\pms{83.7}{1.0} \\
            2-Layer MLP &\pms{83.3}{0.5} \\
            3-Layer MLP & \pms{83.1}{0.5} \\
            \bottomrule
        \end{tabular}}
    \end{minipage}}
    \resizebox{0.35\linewidth}{!}{%
    \begin{minipage}[t]{\pageSplitR\linewidth}
        \centering
        \subfloat[Negatives.\label{tab:expts:c_ablations:n_negatives}]
        {\begin{tabular}{lc}
            \toprule
            & \bf AUCCESS \\
            \midrule
            3 & \pms{82.7}{1.2}   \\
            7 & \pms{83.3}{1.0}  \\
            15 & \pms{82.9}{1.0} \\
            31 & \pms{83.0}{1.0} \\
            \underline{63} & \pms{83.7}{1.0} \\
            \bottomrule
        \end{tabular}}
        \hfill
        \subfloat[Temperature $\beta$. \label{tab:expts:c_ablations:temp}]{\begin{tabular}{lc}
            \toprule
            \bf Temperature & \bf AUCCESS \\
            \midrule
            0.01  & \pms{83.4}{0.5} \\
            \underline{0.1}  & \pms{84.7}{0.8} \\
            1 & \pms{83.7}{1.0} \\
            10 & \pms{82.4}{1.3} \\
            \arrayrulecolor{black}
            \bottomrule
        \end{tabular}}
        \hfill
        \subfloat[Number of frames $K$. \label{tab:expts:c_ablations:n_frames}]{\begin{tabular}{lc}
            \toprule
            \bf K & \bf AUCCESS \\
            \midrule
            1  & \pms{84.7}{1.3} \\
            \underline{2}  & \pms{85.1}{0.9} \\
            3 & \pms{84.7}{0.8} \\
            \arrayrulecolor{black}
            \bottomrule
        \end{tabular}}
    \end{minipage}}
    
    \caption{{\bf Self-supervised loss ablations.} AUCCESS on PHYRE-1B validation folds in within-template setting.
    \label{tab:expts:c_ablations}}
    \vspace{-2mm}
\end{table}

\noindent\textbf{Training.} 
In both cases, the model is trained to minimize $\mathcal{L} + \mathcal{L}_{aux}$, assigning equal weight to both losses.
We follow~\citet{bakhtin2019phyre} and train the model using mini-batch SGD.
We balance the training batches to contain an equal number of positive and negative task-action pairs. 
To facilitate computation of $\mathcal{L}_{aux}$,
we sample $t$ tasks uniformly at random in a batch. 
For each task, we sample $n$ actions that solve the task and $n$ actions that do not solve the task.
We compute the similarity in hand-crafted case, $v_{a,a'}$, for all $4n^2$ action pairs for a task.
To evaluate $\mathcal{L}_{aux}$, we average over these $4tn^2$ action pairs.
Simultaneously, we average $\mathcal{L}$ over the $2tn$ task-action pairs.
For the self-supervised case, we use similar optimization and batch composition as above. For computing $\mathcal{L}_{aux}$ we use as negatives all other rollout embeddings on the same GPU.
See the appendix for more details on our training procedure and hyperparameters.

\noindent\textbf{Inference.} At inference time, for both our models, the agent scores a set of $A$ randomly selected actions using the scoring function $ \hat{y}_{t,a_i}$. The agent proposes the highest-scoring action as a solution. If that action does not solve the task, the agent submits the subsequent highest-scoring action until the task is solved or until the agent has exhausted its attempts (whichever happens first).

\section{Experiments}
\label{sect:expts}
{\bf \noindent Dataset.}
We test our dynamics-aware deep Q-network (DQN) on the PHYRE benchmark on both tiers (1B and 2B) and both generalization settings: within-template and cross-template. Following~\citet{bakhtin2019phyre}, we use all 10 folds and evaluate on the test splits in our final evaluation; see the results in Table~\ref{tab:expts:sota}. For all ablation studies, we use the 4 folds on the validation splits; see Table~\ref{tab:expts:ablations} and~\ref{tab:expts:c_ablations}.

{\bf \noindent Implementation details.}
We train our models on 100,000 batches with 512 samples per batch. Each batch contains 64 unique tasks with 8 actions per task, such that half of them solve the task (positives) and half of them do not (negatives).
Training is performed using Adam~\cite{Kingma14} with an initial learning rate of $3 \cdot 10^{-4}$ and a cosine learning rate schedule~\cite{loshchilov2016}.
We set $K=20$, we set the maximum possible distance in a PHYRE scene $\alpha=\sqrt{2}$, and we set the dimensionality of $\bp_{s,a}$ to 256.
We train and test all models in both the within-template and the cross-template generalization settings.
Following~\citet{bakhtin2019phyre}, we also study the effect of online updates during inference time in the cross-template setting.

\subsection{Comparison to SOTA and Qualitative Analysis}
Table~\ref{tab:expts:sota} presents the AUCCESS of our best dynamics-aware DQNs, and compares it to results reported in prior work.
The results show the strong performance of our models compared to both dynamics-agnostic~\cite{bakhtin2019phyre} and dynamics-modeling~\cite{girdhar2020forward,qi2021longterm} approaches.
In the cross-template setting our self-supervised model improves over the DQN baseline and the Dec[Joint] forward model and almost matches the performance of RPIN.
The results suggest dynamics-aware DQNs have the potential to improve physical-reasoning models.
Figure~\ref{fig:expts:inclination} illustrates how our dynamics-aware DQN model better eliminates parts of the action space in its predictions.

\subsection{Ablation Study}\label{sec:ablations}
We ablate various components of our architecture for hand-crafted and self-supervised losses in Tables~~\ref{tab:expts:ablations} and~~\ref{tab:expts:c_ablations}.
In each subtable, we vary only one component of the model and keep all the other components to fixed.
The values that we used to produce our final results in Table~\ref{tab:expts:sota} are underlined. %

\subsubsection{Hand-Crafted Dynamics-Aware Loss}
{\bf \noindent Effect of Auxiliary loss.} We evaluate in Table~\ref{tab:expts:ablations:backbones}. 
To control for all other factors we train another network with identical parameters to using only $\mathcal{L}$ which results in a DQN agent and we compare it an agent trained with both $\mathcal{L}$ and $\mathcal{L}+ \mathcal{L}_{aux}$. We see that our method outperforms the DQN baseline in the within-template setup and matches the baseline in the cross-template setup. 

{\bf  \noindent  Effect of projection layer.} In Equation~\ref{eqn:approach:joint_embedding}, we described an MLP to project the embedding from the backbone network. In Table~\ref{tab:expts:ablations:embeddor}, we test replacing it with a linear layer or directly using the backbone embeddings.
We find a that a linear layer works slightly better and adopt it in our final model.

{\bf \noindent Effect of number of bins, $\boldsymbol{K}$.} We evaluate its effect in classifying action similarity values in $\mathcal{L}_{aux}$. The results in Table~\ref{tab:expts:ablations:bins} show that $K=20$ performs well but using fewer bins works fine, too. We also compare the bin-classification approach to a regressor that minimizes mean-squared error (MSE) on the action similarities, but find it performs worse.

{\bf \noindent Effect of combination function.} We compare different combination functions $\odot$ for computing the representations $\bj_{s,a,a'}$. Table~\ref{tab:expts:ablations:combination} presents the results of this comparison. We find that element-wise multiplication works better than concatenation, but combination by bilinear layer works best.

{\bf \noindent Effect of frames considered in action similarity measure.} We evaluate the effect of changing the frames of the simulator roll-out used to compute the action similarity, $v_{a,a'}$. The results of this evaluation in Table~\ref{tab:expts:ablations:time} show that using the entire rollout or the last 10 frames works best, with using the first 10 frames or the last 5 frames trail closely behind. 

\subsubsection{Self-Supervised Dynamics-Aware Loss}
We also ablate components of our self-supervised loss.

{\bf \noindent Action representation.} Since we render actions onto frames before computing the action-scene embedding instead of using FiLM, we evaluate the effect of using that representation in Table~\ref{tab:expts:c_ablations:action_rep}. We find that the representation slightly improves slightly over the FiLM representation.

{\bf \noindent Adding contrastive loss.} We vary the relative weight of the contrastive loss in Table~\ref{tab:expts:c_ablations:weight}. The results show that adding the contrastive loss improves over the baseline DQN model.

{\bf \noindent Other contrastive parameters.} We also ablate the projection layer, number of negatives, temperature ($\beta$) for the loss, and the number of frames ($K$).
We find that the contrastive loss deteriorates performance when no projection layer is used, and a linear projection works best. We also find that our method is relatively sensitive to the temperature hyperparameter, and that there is a positive correlation between performance and number of negative samples.

\section{Conclusion}
We have presented a dynamics-aware DQN model for efficient physical reasoning that better captures object dynamics than baseline DQNs by leveraging self-supervised learning on simulator roll-outs. Our model does not require explicit forward prediction at inference time. Our best models outperform prior work on the challenging PHYRE benchmark.

\bibliography{refs}
\bibliographystyle{icml2021}
\clearpage
\appendix
\begin{figure*}[b!]
    \rotatebox{90}{\small ~~~~~~~~~AUCCESS} \hfill
    \subfloat[Template\label{fig:apndx:aucsbytemp}]{
        \includegraphics[height=1.17in]{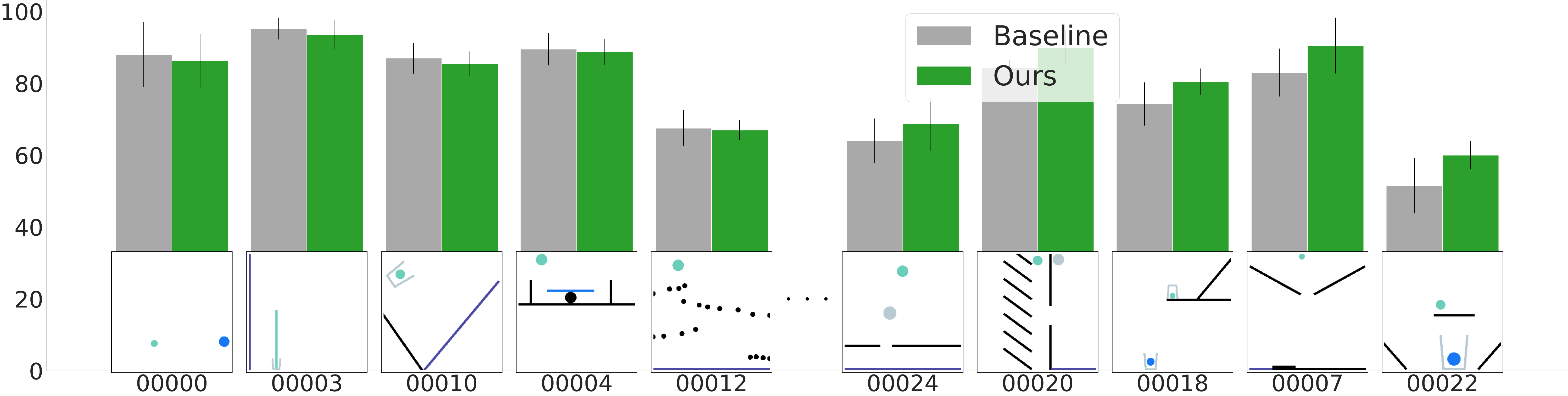}
    }\hfill
    \subfloat[No.\ of moving objects\label{fig:apndx:aucsbymov}]{
        \includegraphics[height=1.17in]{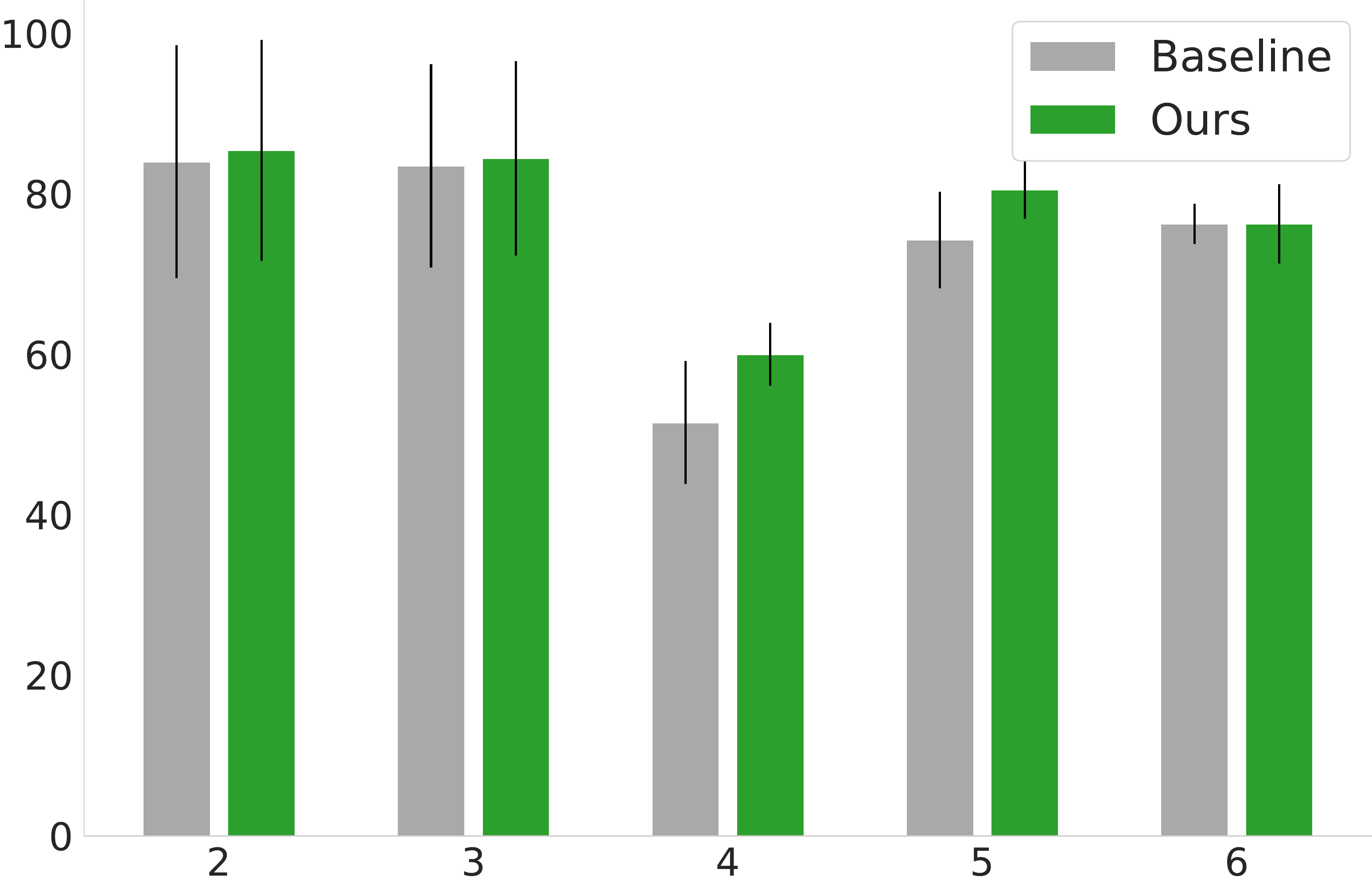}
    }
    \caption{Here we break down the AUCCESS by template in~\ref{fig:apndx:aucsbytemp} and number of moving objects in \ref{fig:apndx:aucsbymov}. We see our agents biggest gains are on templates where the baseline performs worst, while the baseline marginally outperforms our models in the templates where baseline was already performing well. We aggregate the templates by the number of moving objects in where we see our model outperforming the baseline across all numbers of moving objects.}\label{fig:apndx:templateBreakdown}
\end{figure*}

\section{Action Similarity Metric}
\label{apndx:metric}
The similarity between two actions is computed from the object feature representation of the actions' rollouts provided by the PHYRE API. For two rollouts of two actions $(a, a')$ we use the notation that $\left( x(o, t), y(o,t)\right)$ and $\left( x'(o, t), y'(o,t)\right)$ are the locations of the object $o$ at the timestep $t$ in the rollouts of $a$ and $a'$ respectively then:

\begin{align}
T &= \min(t_1, t_2) \\
q_{a,a'}(o, t) &= 1 - \frac{\min(d_{a,a'}(o, t), \alpha)}{\alpha} \label{eqn:apdx:q}\\
v_{a,a'} &= \frac{\sum_{o \in \mathcal{O}} \sum_{t=1}^T q_{a,a'}(o, t)}{T | \mathcal{O} |},
\end{align}

where $\mathcal{O}$ is the set of moving objects in the scene, $t_1$ and $t_2$ are the lengths of the first and second rollouts respectively and $\alpha$ is a hyperparameter that clips the distance at a maximum value.
When computing the metric using only "last $n$" frames the frames we consider are the frames from time $(T-n+1)$ to $T$.
The similarity $ v_{a,a'}$ is binned to construct $\by_s$ as follows:
\begin{align}
    \by_s = \lfloor v_{a,a'} (K - 1) \rceil,
\end{align}
where $\lfloor\cdot\rceil$ is an operator that rounds continuous numbers to the nearest integer and $K$ is the number of bins used.

In Table~\ref{tab:expts:sota}, we take $\alpha=\sqrt{2}$. This value is suggested by the PHYRE environment since the coordinates of locations in the scene fall in the square limited by the corner points $(0,0)$ and $(1,1)$ with the maximum possible Euclidean distance between two objects being $\sqrt{2}$. %

\section{Qualitative Study}
\begin{figure*}[t]
    \centering
    \setlength\tabcolsep{1.5pt}
    \begin{minipage}[t]{0.37\linewidth}
        \centering
        \tiny
        \begin{tabular}{cc}
            \rotatebox{90}{~~~~~~~~~$\Delta$ AUCCESS} &
            \includegraphics[width=0.9\linewidth]{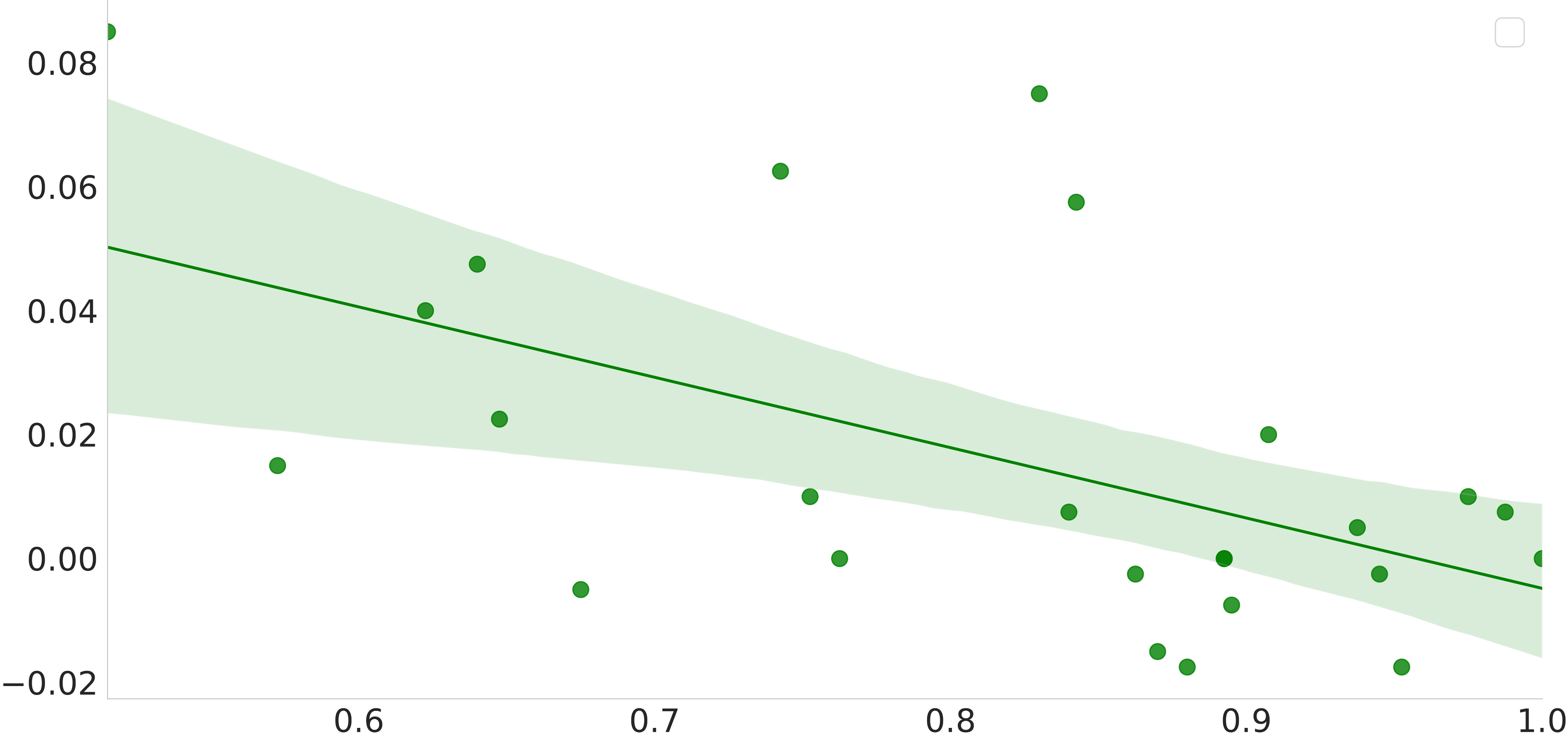} \\
            & Baseline AUCCESS \\
        \end{tabular}
        \captionof{figure}{Here we show baseline AUCCESS vs $\Delta$AUCCESS from our method and find a statistically significant correlation with a Pearson correlation factor of -0.54. This shows we get highest gains on templates where the baseline performs poorly.\label{fig:expts:aucsbytemp}
        \label{fig:apndx:deltacorr}}
    \end{minipage}
    \hfill
    \begin{minipage}[t]{0.6\linewidth}
        \centering
        \tiny
        \begin{tabular}{cccccc}
            GT & & sim$>0$ & sim$>0.5$ & sim$>0.9$ & sim$>0.98$ \\
            \frame{\includegraphics[width=0.18\linewidth]{visualizations/00022660/00022660gt.png}} &
            \rotatebox{90}{~~~~Baseline} &
            \frame{\includegraphics[width=0.18\linewidth]{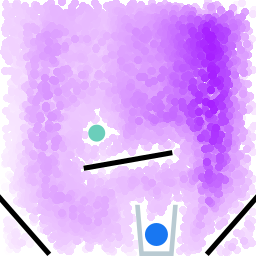}} &
            \frame{\includegraphics[width=0.18\linewidth]{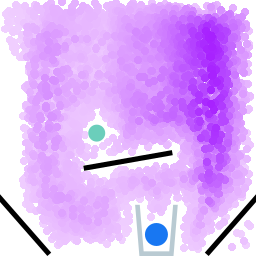}} &
            \frame{\includegraphics[width=0.18\linewidth]{visualizations/action_embedding_sim/threshold_0.90/baseline.png}} &
            \frame{\includegraphics[width=0.18\linewidth]{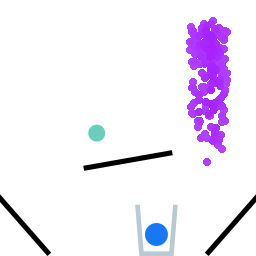}} \\
            &
            \rotatebox{90}{~~~~~Ours} &
            \frame{\includegraphics[width=0.18\linewidth]{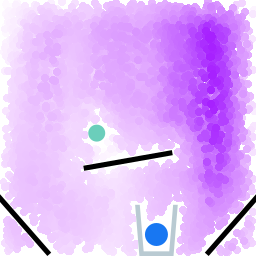}} &
            \frame{\includegraphics[width=0.18\linewidth]{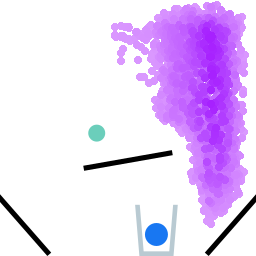}} &
            \frame{\includegraphics[width=0.18\linewidth]{visualizations/action_embedding_sim/threshold_0.90/ours.png}} &
            \frame{\includegraphics[width=0.18\linewidth]{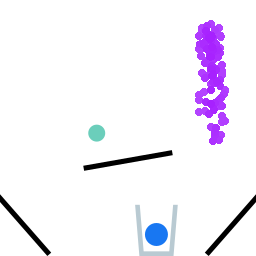}} \\
        \end{tabular}
        \captionof{figure}{Here we show an extended version of Figure~\ref{fig:expts:inclination} (c), showing action space embeddings color coded by similarity to GT actions, at different similarity thresholds. We observe our method leads to actions is able to rule out actions incapable of solving the task at all thresholds, and at 0.98 the selected actions are almost indistinguishable from GT.}\label{fig:appdx:embeddings_full}
    \end{minipage}
\end{figure*}

We break down the AUCCESS improvement by template in Figure~\ref{fig:apndx:templateBreakdown} and try to characterize the the templates in which our method improves most over the baseline and we find in Figure~\ref{fig:apndx:deltacorr} that our method introduces the highest gains in templates which the performance of the baseline was lower. This holds even when comparing across templates where the baseline has still not reached maximum performance.
In Figure~\ref{fig:appdx:embeddings_full}, we show all actions for the given task, color coded by their similarity to GT actions for ours and baseline model. We find our dynamics-aware model is able to rule out incorrect actions much more effectively than the baseline, at all different levels of similarity thresholds considered.

\FloatBarrier

\end{document}